\documentclass{article}

\usepackage[preprint]{neurips_2026}
\makeatletter
\renewcommand{\@notice}{}
\g@addto@macro\@bottomtitlebar{\vskip-11.751pt}
\makeatother
\usepackage[utf8]{inputenc}
\usepackage[T1]{fontenc}
\usepackage{url}
\usepackage{booktabs}
\usepackage{tabularx}
\usepackage{amsmath,amssymb}
\usepackage{graphicx}
\usepackage{wrapfig}
\usepackage{float}
\newcommand{\modellogo}[1]{\raisebox{-0.35em}{\includegraphics[height=14pt]{figures/model-logos/#1.pdf}}}
\DeclareRobustCommand{\inlinemodellogo}[1]{\raisebox{-\dp\strutbox}{\includegraphics[height=\dimexpr\ht\strutbox+\dp\strutbox\relax]{figures/model-logos/#1.pdf}}}
\DeclareRobustCommand{\inlinemethodbadge}[1]{\raisebox{-\dp\strutbox}{\includegraphics[height=\dimexpr\ht\strutbox+\dp\strutbox\relax]{figures/model-logos/badge-#1.pdf}}}
\usepackage{needspace}
\usepackage{microtype}
\usepackage{xcolor}
\usepackage{colortbl}
\usepackage{algorithm}
\usepackage{algpseudocode}
\usepackage{enumitem}
\usepackage{hyperref}

\usepackage[most]{tcolorbox}

\definecolor{promptGray}{HTML}{7B8799}
\DeclareTColorBox{promptbox}{O{Prompt} O{}}{%
	enhanced,
	colback=promptGray!9!white,
	colframe=promptGray,
	coltext=black,
	boxrule=0.7pt, arc=7pt,
	left=9pt, right=9pt, top=9pt, bottom=7pt,
	before skip=13pt, after skip=10pt,
	fontupper=\small,
	title={#1},
	fonttitle=\normalsize\bfseries,
	coltitle=white,
	attach boxed title to top left={xshift=8pt,yshift=-6pt},
	boxed title style={colback=promptGray,colframe=promptGray,boxrule=0pt,
		arc=1.5pt,boxsep=1pt,left=5pt,right=5pt,top=1.5pt,bottom=1.5pt},
	#2
}
\definecolor{findingEvaluator}{HTML}{4D6BF3}
\colorlet{findingColor}{findingEvaluator}
\newtcolorbox[auto counter]{findingbox}[1][]{%
	enhanced,
	colback=findingColor!9!white,
	colframe=findingColor,
	coltext=black,
	boxrule=0.7pt,
	arc=7pt,
	left=9pt, right=9pt, top=9pt, bottom=7pt,
	before skip=13pt, after skip=10pt,
	fontupper=\bfseries,
	title={Finding~\thetcbcounter},
	fonttitle=\normalsize\bfseries,
	coltitle=white,
	attach boxed title to top left={xshift=8pt,yshift=-6pt},
	boxed title style={colback=findingColor,colframe=findingColor,boxrule=0pt,
		arc=1.5pt,boxsep=1pt,left=5pt,right=5pt,top=1.5pt,bottom=1.5pt},
	#1
}

\newtcolorbox{contributionsbox}[1][Main Contributions]{%
	enhanced, colback=findingEvaluator!9!white, colframe=findingEvaluator, coltext=black,
	boxrule=0.7pt, arc=7pt, left=9pt, right=9pt, top=9pt, bottom=7pt,
	before skip=8pt, after skip=6pt,
	title={#1}, fonttitle=\normalsize\bfseries, coltitle=white,
	attach boxed title to top left={xshift=8pt,yshift=-6pt},
	boxed title style={colback=findingEvaluator,colframe=findingEvaluator,boxrule=0pt,
		arc=1.5pt,boxsep=1pt,left=5pt,right=5pt,top=1.5pt,bottom=1.5pt},
}

\title{Actor, Judge, or Teacher? Understanding Decision Models in Long-Horizon Agents}
\author{%
    Haochen Liu\textsuperscript{1,*}, Qianhao Zhu\textsuperscript{1,*}, Zixuan Wang\textsuperscript{1,*}, Dingming Li\textsuperscript{1}, Tongbo Chen\textsuperscript{1} \\
    \textbf{Peng Qiu\textsuperscript{2}, Yingqi Qu\textsuperscript{2}, Weiming Lu\textsuperscript{1}, Jun Xiao\textsuperscript{1}, Yongliang Shen\textsuperscript{1,$\dagger$}} \\
    \textnormal{\textsuperscript{1}Zhejiang University \qquad \textsuperscript{2}Baidu Inc.}
}

\begin{document}
\maketitle
\renewcommand{\thefootnote}{\fnsymbol{footnote}}
\footnotetext[1]{~The first three authors have equal contributions. $^\dagger$~Corresponding author.}
\renewcommand{\thefootnote}{\arabic{footnote}}

\begin{abstract}
Decision models answer a question with a typed output, such as a choice among options with a probability for each, so one model can choose an agent's actions, evaluate actions already taken, or score every action to teach a policy with reinforcement learning. Prior work has studied them mainly as actors, judging their usefulness by how well they act. We ask whether a decision model that acts poorly can still teach well. We compare three decision models and three general LLMs as actors and judges on ALFWorld, and use JEV as a teacher on ALFWorld and WebShop. As actors, all six models are pulled away from a subgoal that must come first by an easier one. As judges that see the whole trajectory, the same models credit the steps toward that subgoal, and asking whether each action helped, rather than whether the task will succeed, separates useful from useless actions. As a teacher, JEV raises a Qwen2.5-1.5B policy over GRPO from 72.8\% to 82.9\% success on ALFWorld and from 56.8\% to 71.4\% on WebShop. These results suggest that a poor actor can still teach well, because judging an action after the episode ends reveals the consequences that acting must anticipate. Code is available at \url{https://github.com/ZJU-REAL/decision-agents}.
\end{abstract}

\section{Introduction}

\label{sec:introduction}

\raggedbottom

Decision models are pretrained models that answer a question about a context with a \emph{typed} output, such as a choice among given options with a probability for each, instead of generated text, and they do so quickly and cheaply~\citep{typesafe2026jevlaunch,openai2026decisions,perplexity2026deciderv12}. An LLM agent can therefore query the same decision model in two ways: to choose its next action, or to evaluate an action that has already been taken. Its evaluations can in turn teach a policy during reinforcement learning, by scoring each action of every rollout. This use matters because outcome rewards give every action in a trajectory the same credit~\citep{NEURIPS2019_195f1538,shao2024deepseekmathpushinglimitsmathematical}, while step-level judgments are otherwise costly, requiring labels for each environment~\citep{ICLR2024_aca97732} or an LLM judge queried at every action~\citep{tan2026hcapo,zhu2026lvrl}. Yet decision models have been evaluated mainly as actors~\citep{zhu2026jevagentharnessbenchmarkingdeployingtraining,ling2026fasterdecisions}, and in that role they often fail: they are drawn to commands that mention task objects, skip prerequisite subgoals, and complete subgoals in the wrong order~\citep{yang2026codeownssimulationjev,li2026reading,ling2026fasterdecisions}.

These failures do not settle how useful a decision model is when it evaluates rather than acts. On the same ALFWorld evaluation tasks, JEV completes 42.9\% of the tasks when it chooses the actions itself. When it instead scores every action of a 1.5B policy during training, after seeing the whole trajectory, the policy completes 82.9\% of the tasks if JEV is asked whether each action helped achieve the goal, compared with 72.8\% for training with outcome rewards alone. Asked instead whether the task will succeed after each action, JEV yields a policy that completes only 69.3\%. The same model can thus act poorly yet teach well, but only when it is asked the right question. This paper asks \textbf{why a decision model that acts poorly can still teach well, and which queries make this possible}.

\begin{figure}
	\centering
	\setlength{\abovecaptionskip}{2pt}
	\includegraphics[width=\linewidth,trim=0bp 0bp 118bp 0bp,clip]{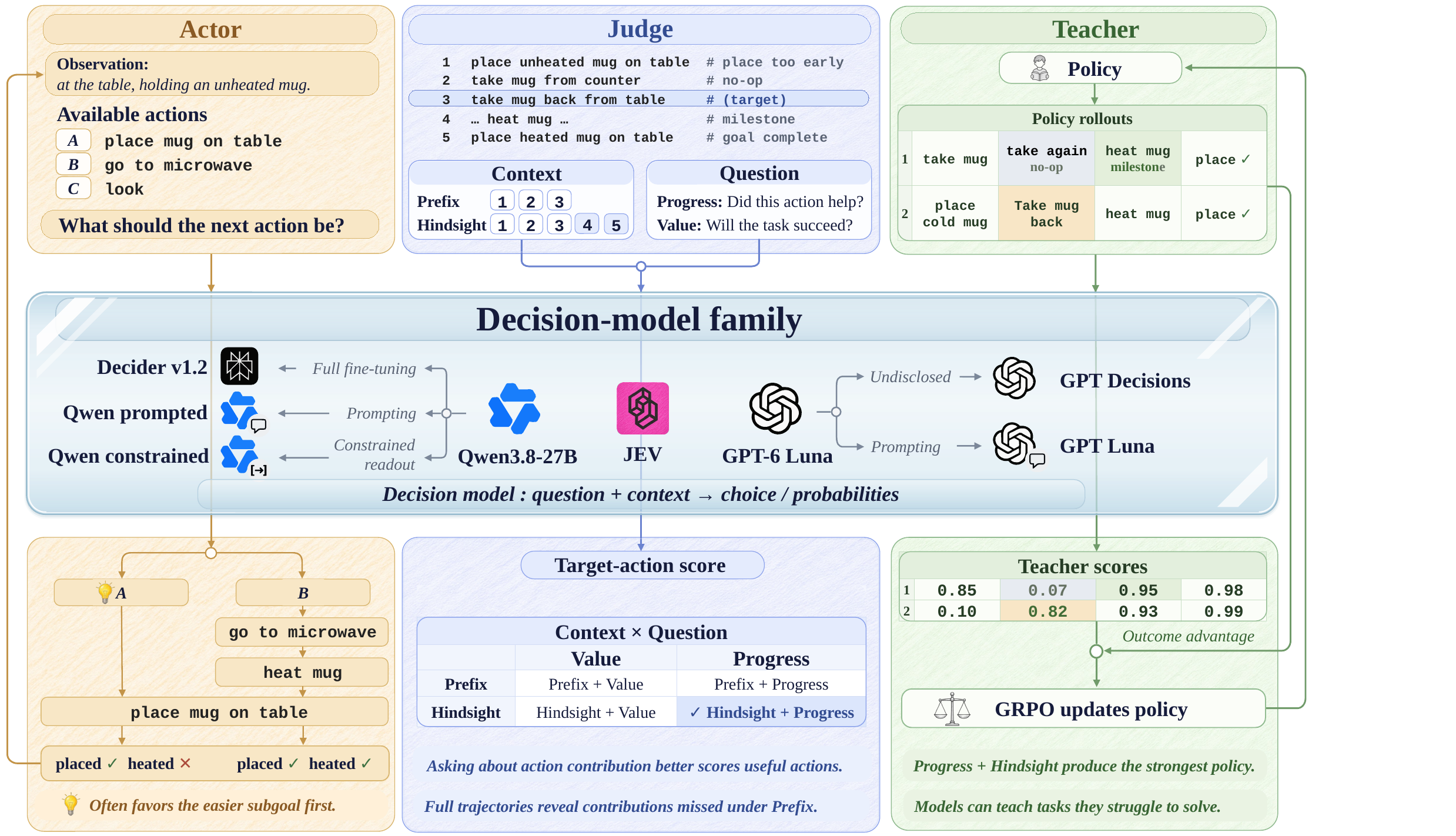}\par
	\caption{One decision model queried in three ways, on the task of putting a heated mug on a table. \textbf{Actor}: choosing the next action. \textbf{Judge}: scoring an action of a recorded trajectory under two contexts (Prefix, Hindsight) and two questions (Value, Progress). \textbf{Teacher}: scoring the actions of policy rollouts, with the centered scores added to the GRPO outcome advantage. The middle row lists the six models we evaluate.}
	\label{fig:jev-overview}
	\vspace{-8pt}
\end{figure}

We answer these questions by querying the same decision models in each way on the same tasks (Figure~\ref{fig:jev-overview}). As an \textbf{Actor}, a model chooses each action before seeing its effect. As a \textbf{Judge}, it evaluates actions in recorded trajectories, and we vary two choices: what it sees, either the trajectory up to the action (\emph{Prefix}) or the whole trajectory (\emph{Hindsight}), and what it is asked, either whether the task will succeed after the action (\emph{Value}) or whether the action helped achieve the goal (\emph{Progress}). As a \textbf{Teacher}, it applies the same queries to the actions of a policy being trained, and its scores, centered within each rollout group, are added to the outcome advantage of GRPO~\citep{shao2024deepseekmathpushinglimitsmathematical}. Each analysis changes one factor, such as the goal, the question, or the context, and holds the rest fixed. We evaluate three decision models and three general LLMs given the same questions, two of them the base LLMs of decision models, on ALFWorld~\citep{shridhar2021alfworld} and WebShop~\citep{NEURIPS2022_82ad13ec}, where some subgoals must be completed in order.

\paragraph{Acting fails when an easier subgoal draws the model away from the required one.}

All six models succeed far less often when one subgoal must be completed before another, such as heating a mug before placing it on a table; JEV, for example, completes 70.8\% of tasks with two independent subgoals but only 14.7\% of tasks with ordered ones. The models do not lack knowledge of the required step. In states where a single action would complete the easier subgoal, removing that subgoal from the goal moves most of each model's probability to the required step; for JEV, the probability on the required state change rises from 14.6\% to 77.6\%. A reminder about the dependency changes success by only $-2.8\%$ to $+5.2\%$. To act correctly, the model must anticipate that the easier action leaves the task incomplete before any consequence is visible, whereas a judge evaluates the action after its consequences have occurred.

\paragraph{Whether a judge avoids this failure depends on what it sees and what it is asked.}

Seeing the whole trajectory helps Judges recognize contributions that the same models overlook as Actors: these contributions can be unclear from the prefix alone but become apparent from what follows. Seeing the consequences is not enough, however. With the whole trajectory visible, asking whether each action helped separates actions that complete a subgoal from actions that change nothing, with JEV scoring the former 0.36 higher within a trajectory, whereas asking whether the task will succeed gives a gap of only 0.05, because every action in the trajectory then has the same answer, the outcome. All six models show both effects, and the two decision models with a disclosed base LLM judge more accurately than that LLM given the same question, at 28\% to 43\% lower cost per score.

\paragraph{The same choices determine how much a decision model teaches.}

The question and context that judge best also teach best. The ALFWorld result above holds on WebShop, where JEV's Progress scores with the whole trajectory raise success from 56.8\% with outcome rewards alone to 71.4\%, against 68.3\% with the Value question, and the 1.5B policy taught this way outperforms a 7B policy trained with outcome rewards alone on both environments. With only the prefix visible, the two questions perform similarly. The gains are largest on the ALFWorld tasks with ordered subgoals, where the policy completes 86.8\%, 6.9\% to 16.2\% more than with the other questions and contexts, although JEV completes only 14.7\% of these tasks as an actor.

Taken together, these results suggest that \emph{a decision model that acts poorly can still teach well, because judging an action after the episode ends reveals the consequences that acting must anticipate}. This benefit requires asking about each action rather than about the outcome, and it is largest for the actions the model handles worst as an actor, those whose value depends on what follows. Two implications follow for using decision models, and possibly other frozen judges, in agent training. First, a model's suitability as a supervisor should be assessed by how well it judges recorded trajectories, not only by how well it acts. Second, step-level supervision can be obtained cheaply by querying a frozen decision model after each episode, without step labels or a learned critic, for 22 to 30 USD of queries per training run in our experiments.

\begingroup
\setlength{\intextsep}{6pt}
\begin{table}[H]
\centering
\caption{Models evaluated in this work. \emph{Adaptation} describes how a decision model is obtained from its base LLM, and \emph{Access} indicates whether the model weights are released.}
\label{tab:models}
\small
\renewcommand{\modellogo}[1]{\raisebox{-0.35em}{\includegraphics[height=12pt]{figures/model-logos/#1.pdf}}}
\setlength{\tabcolsep}{8pt}
\begin{tabular*}{\textwidth}{@{\extracolsep{\fill}}llll@{}}
\toprule
Model & Base LLM & Adaptation & Access \\
\midrule
\rowcolor{gray!12}
\multicolumn{4}{c}{\textit{Dedicated decision models}} \\
\modellogo{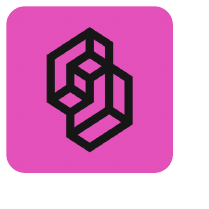}\, JEV & -- & -- & API only \\
\modellogo{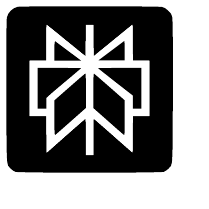}\, Decider v1.2 & Qwen3.8-27B & Full fine-tuning & Open weights \\
\modellogo{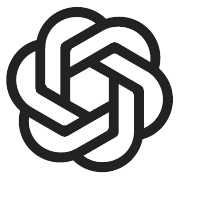}\, GPT Decisions & GPT-6 Luna & -- & API only \\
\addlinespace[2pt]
\rowcolor{gray!12}
\multicolumn{4}{c}{\textit{General LLMs as decision models}} \\
\modellogo{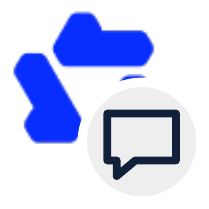}\, Qwen prompted & Qwen3.8-27B & Prompting & Open weights \\
\modellogo{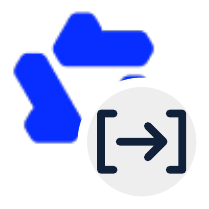}\, Qwen constrained & Qwen3.8-27B & Constrained readout & Open weights \\
\modellogo{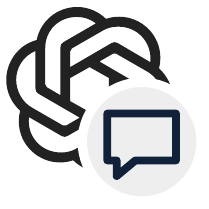}\, GPT Luna & GPT-6 Luna & Prompting & API only \\
\bottomrule
\end{tabular*}
\end{table}
\endgroup

\section{Experimental Setup}
\label{sec:analysis_setup}

\subsection{Decision Models}
\label{sec:decision-models}

A decision model takes a question and a context as input, and its output depends on the question type: for a multiple-choice question, it selects an option and assigns a probability to every option; for a yes/no question, it returns a single score in $[0,1]$. We evaluate six models that answer questions in this form, grouped by how the decision interface is obtained (Table~\ref{tab:models}).

\paragraph{Dedicated decision models.}
These models are released by their providers specifically as decision models and are queried through their own decision APIs. JEV~(\inlinemodellogo{jev})~\citep{typesafe2026jev} is the earliest public decision model; it is closed, and neither its base LLM nor its adaptation procedure is disclosed. Decider v1.2~(\inlinemodellogo{decider})~\citep{perplexity2026deciderv12} has open weights and is fully fine-tuned from Qwen3.8-27B~\citep{qwen38}. GPT Decisions~(\inlinemodellogo{gpt-decisions})~\citep{openai2026decisions} is closed and built on GPT-6 Luna, with an undisclosed adaptation procedure.

\paragraph{General LLMs as decision models.}
A general LLM can answer the same questions when queried in a suitable way, with its weights and architecture unchanged. We apply two common methods to the base LLMs of Decider v1.2 and GPT Decisions. \textit{Prompting}~(\inlinemethodbadge{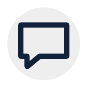}) instructs the model to output its choice and probabilities in JSON, which we then parse. \textit{Constrained readout}~(\inlinemethodbadge{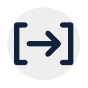}) uses the first-token probabilities returned for valid answer labels. For multiple-choice questions, we select the highest-scoring returned option label and normalize the probabilities over the returned option labels. For yes/no questions, we normalize over Yes and No to obtain the score. Applying both methods to Qwen3.8-27B gives Qwen prompted~(\inlinemodellogo{qwen-prompted}) and Qwen constrained~(\inlinemodellogo{qwen-constrained}), and applying prompting to GPT-6 Luna gives GPT Luna~(\inlinemodellogo{gpt-prompted}). Reasoning is disabled for all three. JEV has no such counterpart because its base LLM is undisclosed.

Comparing each dedicated decision model with its own corresponding LLM queried in these ways shows what the dedicated model adds. For Decider v1.2, the adaptation is fine-tuning; for GPT Decisions, the adaptation procedure has not been disclosed. Appendix~\ref{app:prompts} gives the model identifiers and interface details.

\subsection{Benchmarks}
\label{sec:benchmarks}

We use two text-based environments in which a task contains several requirements, some of which must be satisfied in a fixed order.

\vspace{-3pt}
\paragraph{ALFWorld.}
ALFWorld~\citep{shridhar2021alfworld} is a household environment in which an agent completes goals such as ``put a clean sponge on the cart'' by issuing text commands such as \texttt{go to}, \texttt{take}, and \texttt{clean}. Its six task types fall into three groups according to how their requirements relate. \textit{Pick} and \textit{Look} involve a single object. \textit{Clean}, \textit{Heat}, and \textit{Cool} are tasks with dependencies: satisfying the placement requirement depends on first completing the required change to the object's state. \textit{Pick2} also has two requirements, one for each object, but they can be satisfied in either order. Comparing \textit{Pick2} with \textit{Clean}, \textit{Heat}, and \textit{Cool} lets us examine how models handle requirements with an ordering dependency (Section~\ref{sec:jev-actor}).

\vspace{-3pt}
\paragraph{WebShop.}
WebShop~\citep{NEURIPS2022_82ad13ec} is a shopping environment in which an agent searches a product catalogue, opens product pages, selects options such as size and color, and purchases an item. Each instruction specifies the product attributes, options, and a price limit. The required options must be selected before purchase. The final reward measures how well the purchased item matches the instruction, and a task succeeds only when the reward is full.

\vspace{-2pt}
Appendix~\ref{app:setup} gives the training and evaluation splits, inference seeds, and action budgets.

\vspace{-4pt}
\section{Understanding Decision Models on Agentic Tasks}
\label{sec:analysis}

We study each decision model in three roles that differ in what the model is asked and how its answers are used (see Figure~\ref{fig:jev-overview} for an overview and Appendix~\ref{app:prompts} for the prompt templates).
\begin{contributionsbox}[Roles of Decision Models]
\setlength{\leftmargini}{1.2em}
\begin{itemize}[topsep=0pt,itemsep=1pt,parsep=0pt]
		\item \textbf{Actor.} At each step, the decision model answers a multiple-choice question about which action to take next. It receives the goal, the interaction history, and the current observation, then selects one of the valid actions and assigns a probability to each option. The selected action is executed in the environment (Section~\ref{sec:jev-actor}).
		\item \textbf{Judge.} The decision model scores the actions in trajectories recorded from other policies, as in LLM-as-a-judge~\citep{NEURIPS2023_91f18a12}. For each executed action, it answers a yes/no question about that action, such as whether it helped achieve the goal, and returns a score in $[0,1]$ (Section~\ref{sec:judge}).
		\item \textbf{Teacher.} The decision model scores actions as a Judge does, but on trajectories from the policy being trained. The scores are combined with outcome rewards in GRPO~\citep{shao2024deepseekmathpushinglimitsmathematical} as step-level supervision. The decision model stays fixed during training and is not used at test time (Section~\ref{sec:teacher}).
\end{itemize}
\end{contributionsbox}
\vspace{0pt}
\par

\begin{table}[t]
\centering
\caption{Actor success rates (\%) on ALFWorld, using the same evaluation tasks as the trained policies (Section~\ref{sec:teacher}). Each Actor is evaluated once per task. Bold marks the best result in each column. Appendix~\ref{app:actor-performance-l1} reports results on broader validation tasks.}
\label{tab:actor-performance}
\small
\begin{tabular*}{\textwidth}{@{\extracolsep{\fill}}lccccccc@{}}
\toprule
	& \multicolumn{3}{c}{No dependency} & \multicolumn{3}{c}{With dependency} & \\
\cmidrule(lr){2-4} \cmidrule(lr){5-7}
Actor & Pick & Look & Pick2 & Clean & Heat & Cool & All \\
\midrule
\multicolumn{8}{l}{\textit{Dedicated decision models}} \\
\modellogo{jev}\, JEV              & 77.1 & 46.2 & 70.8 & 11.1 & 18.8 & 16.0 & 42.9 \\
\modellogo{decider}\, Decider v1.2     & \textbf{97.1} & 53.8 & \textbf{83.3} & \textbf{55.6} & 18.8 & \textbf{40.0} & \textbf{63.6} \\
\modellogo{gpt-decisions}\, GPT Decisions    & 88.6 & \textbf{76.9} & 62.5 & 29.6 & 25.0 & 24.0 & 52.9 \\
\midrule
\multicolumn{8}{l}{\textit{General LLMs as decision models}} \\
\modellogo{qwen-prompted}\, Qwen prompted    & 77.1 & 53.8 & 79.2 & 22.2 & \textbf{31.2} & 20.0 & 49.3 \\
\modellogo{qwen-constrained}\, Qwen constrained & 77.1 & 46.2 & 75.0 & 33.3 & 6.2 & 8.0 & 45.0 \\
\modellogo{gpt-prompted}\, GPT Luna         & 77.1 & 46.2 & 37.5 & 18.5 & 0.0 & 4.0 & 34.3 \\
\bottomrule
\end{tabular*}
\vspace{-12pt}
\end{table}

\subsection{Decision Models as Actors}
\label{sec:jev-actor}

\begin{findingbox}[label={find:actor},top=7pt,bottom=5pt,before skip=10pt,after skip=6pt]
As Actors, decision models struggle with dependent subgoals: an easier subgoal draws them away from the subgoal that must be met first.
\end{findingbox}

\begin{wrapfigure}[13]{r}{0.5\textwidth}
	\vspace{\dimexpr-\intextsep+13.6pt\relax}
	\centering
	\setlength{\abovecaptionskip}{2.65pt}
	\includegraphics[width=\linewidth]{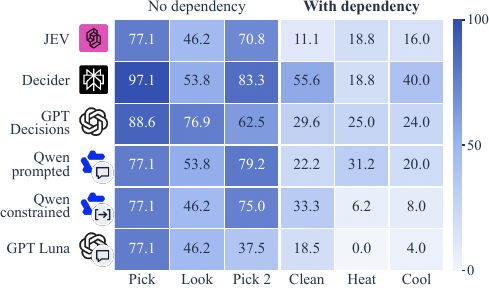}\par
\caption{Actor success rates (\%) on ALFWorld.}  \label{fig:actor-task-heatmap}
\end{wrapfigure}

Table~\ref{tab:actor-performance} and Figure~\ref{fig:actor-task-heatmap} report Actor success rates by ALFWorld task type. In \textit{Clean}, \textit{Heat}, and \textit{Cool} tasks, the actor must clean, heat, or cool an object and place it at a destination; we refer to them as \textbf{tasks with dependency}. All six models succeed less often on these tasks than on \textit{Pick} and \textit{Pick2}. For example, JEV~(\inlinemodellogo{jev}) completes 70.8\% of \textit{Pick2} tasks but at most 18.8\% of tasks with dependency in each state-change type. Decider v1.2~(\inlinemodellogo{decider}), the strongest Actor overall (63.6\%), succeeds on only 18.8\% of \textit{Heat} tasks.

\Needspace{18\baselineskip}

Tasks with dependency and \textit{Pick2} tasks both have two subgoals. In ``put a heated mug on the side table,'' the mug must be heated and on the side table. In ``put two pencils in the drawer,'' both pencils must be in the drawer. The pencils can be placed in either order, but the mug must be heated before placement. Higher success on \textit{Pick2} suggests it is ordering that makes these tasks difficult.

Failed trajectories show how this order is broken. Suppose the actor holds the mug at the side table. It can place the mug immediately in one action. Alternatively, it can go to the microwave, heat it, and return before placing it. In many failures, the actor places the unheated mug directly, leaving the task incomplete.
We next ask whether the easier subgoal draws actors away from the subgoal that must be met first but takes steps to accomplish (Section~\ref{sec:actor-goal-conditions}), and whether reminding them of the dependency helps (Section~\ref{sec:actor-condition-dependency}).

\begin{figure}[H]
	\vspace{-4pt}
	\centering
	\setlength{\abovecaptionskip}{2pt}
	\includegraphics[width=\linewidth]{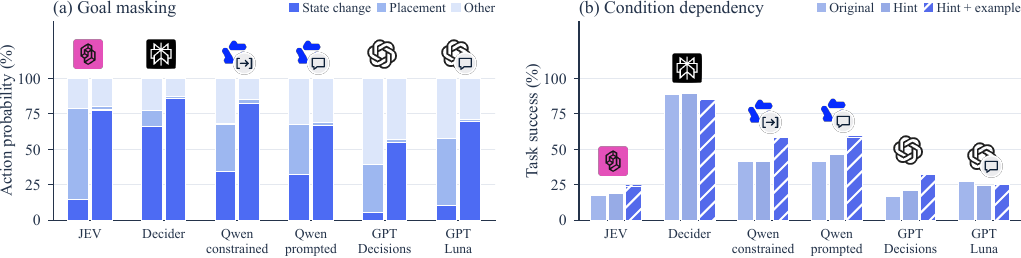}\par
	\caption{(a) Action probabilities with (left) and without (right) the placement subgoal. (b) Success rates with the original prompt, a dependency hint, and a hint plus heated-mug example.}
	\label{fig:actor-analysis}
\end{figure}

\subsubsection{Does the Easier Subgoal Distract Actors?}
\label{sec:actor-goal-conditions}

We collect states in which the actor could satisfy the easier subgoal with a single action, while the subgoal that must come first is still unmet. In ALFWorld, the actor stands at the destination holding an object that still needs to be cleaned, heated, or cooled. This yields 116 states from 66 tasks, and every actor is queried at every state. At each state, we query the actor under two goals: the full goal and the same goal without the easier subgoal (e.g., ``put a heated mug on the side table'' versus ``heat a mug''). The history, observation, candidate actions, and remaining budget are identical in both queries. We group the probabilities assigned to candidate actions into three categories: actions toward the subgoal that must come first, the action that satisfies the easier subgoal, and other actions.

\begingroup
\setlength{\parskip}{5.5pt}
Figure~\ref{fig:actor-analysis}a shows the results. With the full goal, JEV~(\inlinemodellogo{jev}) assigns 63.8\% of its probability to the easier subgoal and only 14.6\% to the subgoal that must come first. Without the easier subgoal, these become 2.5\% and 77.6\%. All six actors shift in the same direction. The actors can therefore identify the required step when it is the only subgoal, but adding an easier subgoal to the goal draws probability away from it.

\subsubsection{Can a Hint Prevent This Error?}
\label{sec:actor-condition-dependency}

We next test whether a hint helps actors avoid this error. We compare three prompts on the same 116 states with the full goal: the original question, the question with ``Consider possible dependencies among task requirements'' appended, and the hint followed by an ordering example. The example starts with a mug neither heated nor on the table, and explains that it must be heated before being placed on the table. Each continuation uses the same history, environment state, candidate actions, and remaining budget, with its prompt applied at every step.

Figure~\ref{fig:actor-analysis}b compares success rates across the three conditions. The hint alone changes success by $-2.8\%$ to $+5.2\%$ across the six actors. Adding the example raises success from 41.7\% to 58.3\% for Qwen constrained~(\inlinemodellogo{qwen-constrained}), from 46.6\% to 58.6\% for Qwen prompted~(\inlinemodellogo{qwen-prompted}), and from 20.8\% to 32.1\% for GPT Decisions~(\inlinemodellogo{gpt-decisions}). JEV~(\inlinemodellogo{jev}) improves more modestly, from 19.0\% to 24.1\%, while GPT Luna~(\inlinemodellogo{gpt-prompted}) changes little and Decider v1.2~(\inlinemodellogo{decider}) declines from 89.7\% to 85.3\%. An ordering example thus helps some actors, but gains vary across models and substantial failures remain.

\vspace{-3pt}
\subsection{Decision Models as Judges}
\label{sec:judge}

\begin{findingbox}[label={find:judge},top=7pt,bottom=5pt,before skip=10pt,after skip=12pt]
As Judges, the same models can recognize contributions they overlook as Actors. This depends on what they see and what they are asked. Seeing the whole trajectory reveals how earlier actions contribute to later goal completion. Asking whether each action helped achieve the goal then distinguishes their contributions more clearly than asking whether the task will succeed, which gives similar scores to actions within a trajectory.
\end{findingbox}

\begin{wrapfigure}[12]{r}{0.5\textwidth}
    \vspace{\dimexpr-\intextsep-1.1pt\relax}
    \centering
    \setlength{\abovecaptionskip}{-1.824pt}
    \includegraphics[width=\linewidth]{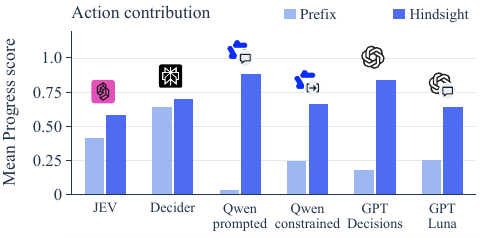}\par
    \caption{Prefix (left) and Hindsight (right) scores on the same states in Section~\ref{sec:actor-goal-conditions}.}
    \label{fig:judge-prefix-full}
\end{wrapfigure}

Section~\ref{sec:jev-actor} shows that Actors struggle when subgoals must be completed in a particular order. However, this leaves open how the same models perform as Judges when they encounter such tasks. We reuse the new trajectories generated under the original prompt in Section~\ref{sec:actor-condition-dependency}. For each successful continuation, we evaluate the first action taken from the corresponding diagnostic state and ask whether this action helped achieve the goal. We compare two choices of \textbf{context} given to the six Judges: the trajectory up to the action (\textit{Prefix}) or the whole trajectory (\textit{Hindsight}), with outcome labels masked. Prefix therefore shows how the models judge an action at the same decision point studied in Section~\ref{sec:actor-goal-conditions}. The model, full goal, target action, and question remain the same.

Figure~\ref{fig:judge-prefix-full} shows higher mean contribution scores with \textit{Hindsight}. JEV's~(\inlinemodellogo{jev}) mean score rises from 0.41 to 0.59. All six models show the same pattern. It seems that when observing full trajectories, decision models can assign higher scores to such actions. But it also raises several questions. Full trajectory may substantially suggest the outcome of the task. Could this lead models to give every action in a trajectory an evenly high score (Section~\ref{sec:judge-question})? What additional information do models gain from seeing the full trajectory (Section~\ref{sec:judge-context})?

\par
\endgroup

\begingroup
\setlength{\parskip}{5.5pt}
\vspace{-4pt}
\subsubsection{Can Judges Distinguish Different Action Contributions?}
\label{sec:judge-question}

As Actors, decision models choose actions to maximize the chance of completing the task. When using the same models as Judges, we can retain this focus on task success and ask whether the task will succeed after the action (\emph{Value}), or ask whether the action helped achieve the goal (\emph{Progress}).
We compare these two questions using the same Hindsight context to test whether shifting from predicting task success, as in the Actor role, to evaluating action contribution changes how Judges score individual actions.

\begin{figure}[t]
	\vspace{-4pt}
	\centering
	\setlength{\abovecaptionskip}{2pt}
	\includegraphics[width=\linewidth]{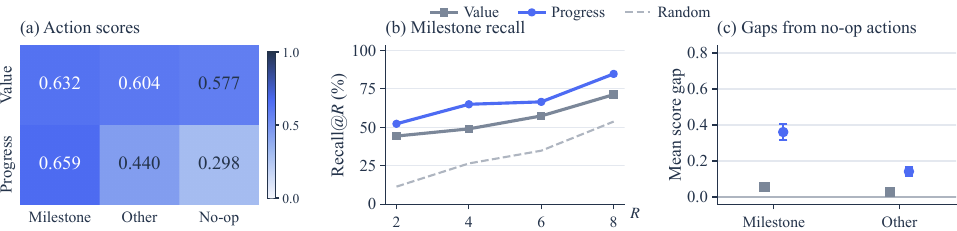}\par
\caption{JEV~(\inlinemodellogo{jev}) as a Judge with Hindsight context on ALFWorld. (a) Mean action scores. (b) Milestone Recall@$R$, grouped by the number of milestone actions $R$ in a trajectory. (c) Within-trajectory score gaps relative to no-op actions.}
	\label{fig:jev-evaluator-objective}
	\vspace{-4pt}
\end{figure}

To address this problem, we group actions into three categories. A \textbf{milestone} action achieves a subgoal~\citep{pmlr-v306-wang26jm}. A \textbf{no-op} action leaves the environment and observation unchanged. Actions of neither type are grouped as \textbf{other} actions. Appendix~\ref{app:evaluator-protocols} describes the labeling procedure. We compare the two questions on 462 ALFWorld trajectories, each containing at least one milestone and one no-op action.

Figure~\ref{fig:jev-evaluator-objective}a shows JEV's~(\inlinemodellogo{jev}) mean scores. Under Value, milestone, other, and no-op actions receive similar scores (0.632, 0.604, and 0.577). Under Progress, the scores spread to 0.659, 0.440, and 0.298. All six Judges show a larger gap between milestone and no-op actions under Progress (Appendix~\ref{app:judge-heatmaps}).

We next check how many milestone actions a Judge places at the top of each trajectory. For a trajectory with $R$ milestone actions, we take its $R$ highest-scored actions and measure the fraction that are milestones (Recall@$R$). Figure~\ref{fig:jev-evaluator-objective}b groups trajectories by $R$, from simpler tasks with two milestone actions to more complex ones with eight. Progress finds more milestones than Value at every $R$, by 8.0\% to 16.0\%. Within the same trajectory, Progress also scores milestone actions 0.36 higher than no-op actions, whereas Value scores them only 0.05 higher (Figure~\ref{fig:jev-evaluator-objective}c).

Judges do not inherently assign similar scores to actions in the same trajectory. When the question retains the Actor's focus on task success, scores differ little across action types. Asking about each action's contribution produces clearer distinctions, suggesting that the question should be adapted when decision models are used to judge individual actions.

\par
\endgroup

\subsubsection{What Do Judges Gain from Observing Full Trajectories?}
\label{sec:judge-context}

\begin{wrapfigure}[16]{r}{0.49\textwidth}
    \vspace{\dimexpr-\intextsep+0.2pt\relax}
    \centering
    \setlength{\abovecaptionskip}{5.73pt}
    {\small\color[HTML]{24334A}Action Ranking Shifts\par}
    \vspace{-1.2pt}
    \includegraphics[width=\linewidth]{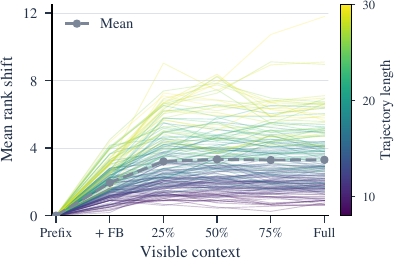}\par
    \makeatletter
    \long\def\@makecaption#1#2{%
        \vskip\abovecaptionskip
        {\raggedright #1: #2\par}%
        \vskip\belowcaptionskip}
    \makeatother
    \caption{JEV's action rankings relative to Prefix. Each thin line is a trajectory; the dashed line is the mean.}
    \label{fig:judge-context}
\end{wrapfigure}

Prefix shows an action together with the history leading to it, whereas Hindsight also reveals how its effects are used in later progress toward the goal. We examine how this additional information changes Judges' assessments.

\vspace{3pt}
We reuse the 167 trajectories from Section~\ref{sec:judge} and evaluate all 2,849 actions. Holding the question fixed to Progress, we progressively extend the visible context. Prefix ends with the target action and excludes the observation returned after its execution. We first add this observation, which we call immediate feedback (FB), and then reveal 25\%, 50\%, 75\%, and all of the remaining trajectory. For each context, we rank actions by their scores within each trajectory and measure the average rank shift, $\mathsf{mean}\,|\mathsf{current\ rank}-\mathsf{Prefix\ rank}|$. We average this quantity across trajectories to show how much additional context reshuffles the relative contributions assigned to actions.

\vspace{3pt}
Figure~\ref{fig:judge-context} shows JEV's~(\inlinemodellogo{jev}) results. Adding immediate feedback shifts actions by an average of 1.95 ranking positions relative to Prefix. This increases to 3.30 positions when the full trajectory is visible. Overall, action rankings move further from their Prefix ordering as more of the trajectory becomes visible. Results for all six Judges are shown in Appendix~\ref{app:judge-context-models}.

\begin{figure}[H]
    \vspace{-5pt}
    \centering
    \setlength{\abovecaptionskip}{2pt}
    \includegraphics[width=\linewidth]{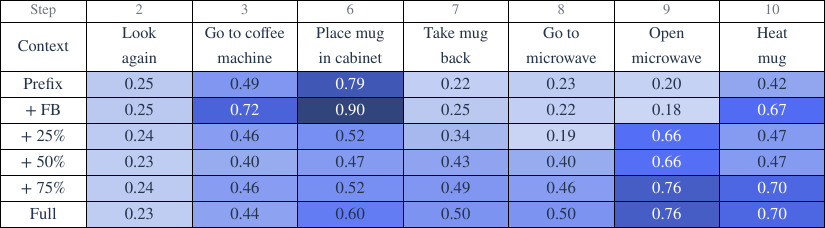}\par
    \caption{JEV's scores for selected steps from a trajectory used in Section~\ref{sec:judge}.}
    \label{fig:judge-context-case}
    \vspace{-13pt}
\end{figure}

Figure~\ref{fig:judge-context-case} illustrates these changes in a task requiring a heated mug to be placed in a cabinet. The mug is placed in the cabinet before heating (6), then taken back (7), heated using the microwave (10), and placed back in the cabinet (12). The complete trajectory is provided in Appendix~\ref{app:context-example}. Across all contexts, JEV~(\inlinemodellogo{jev}) assigns consistently low scores to the repeated look action (2). The preceding look (1) already provides the same observation, allowing this repetition to be judged from Prefix alone. Under Prefix, however, it scores premature placement (6) at 0.79, compared with only 0.23 for going to the microwave (8) and 0.20 for opening it (9). This mirrors the Actor's preference for satisfying the easier subgoal over taking steps toward the prerequisite subgoal in Section~\ref{sec:jev-actor}. Immediate feedback reinforces this pattern: the placement (6) score rises to 0.90, while the scores for going to (8) and opening (9) the microwave fall to 0.22 and 0.18.

\vspace{-3pt}
As more of the full trajectory becomes visible, JEV assigns greater credit to these initially low-scored actions. The score for opening the microwave (9) rises to 0.66 once the visible continuation includes heating the mug (10) and its feedback, and reaches 0.76 with the full trajectory. This example connects the Actor's difficulty with subgoal dependencies to the Judge's use of Hindsight: observing the full trajectory helps JEV recognize the dependency between heating the mug and the preceding steps needed to accomplish it, which is not apparent in its judgments from immediate feedback alone.

\vspace{-4pt}
\subsection{Decision Models as Teachers}
\label{sec:teacher}

\begin{findingbox}[label={find:teacher},top=7pt,bottom=5pt,before skip=8pt,after skip=6pt]
The judgments that best distinguish action contributions also provide the most effective supervision for policy learning. Scoring whether each action helped achieve the goal after seeing the whole trajectory produces the strongest policies, with the largest gains on tasks that the Actor struggles to complete. The trained policies outperform the decision model that teaches them, especially on tasks it rarely completes as an Actor.
\end{findingbox}

Sections~\ref{sec:jev-actor} and~\ref{sec:judge} show that decision models struggle to act on tasks with dependent subgoals, yet can avoid this limitation and can judge action contributions differently when given an appropriate question and the whole trajectory. We next test whether these judgments provide useful supervision for policy learning. We use decision models as Teachers to score actions in trajectories generated by the policy being trained, and combine these scores with outcome rewards as step-level supervision. The decision model stays fixed during training and is not queried at test time.

\vspace{-5pt}
\paragraph{Training with Teacher scores.}
We add the Teacher's scores to GRPO~\citep{shao2024deepseekmathpushinglimitsmathematical} so that actions in the same trajectory can receive different training signals. For each task $u$ with goal $g$, the policy $\pi_\theta$ samples a group $\mathcal{G}_u$ of trajectories. After the rollouts finish, the Teacher $D$ scores each executed action $t$ in each trajectory $\tau_i$:
\begingroup
\setlength{\abovedisplayskip}{3pt}
\setlength{\belowdisplayskip}{3pt}
\setlength{\abovedisplayshortskip}{0pt}
\setlength{\belowdisplayshortskip}{2pt}
\begin{equation}
		s_{i,t} = D(g, \tau_i, t) \in [0,1].
\end{equation}
The input contains the goal and the observations and actions in the trajectory, but no environment rewards, success labels, or hidden states. Under Prefix context, the Teacher sees only the part of $\tau_i$ up to action $t$. By default, the Teacher answers the Progress question with Hindsight context; Section~\ref{sec:teacher-scores} compares all four combinations.
\par
\endgroup

We center the scores within the group and add them to the outcome advantage $A^{\mathrm{out}}_{i,t}$ of GRPO:
\begin{equation}
		A_{i,t} = A^{\mathrm{out}}_{i,t} + \lambda \bigl( s_{i,t} - \bar{s}_{u} \bigr), \qquad
		\bar{s}_{u} = \frac{1}{|\mathcal{S}_u|} \sum_{(j,k) \in \mathcal{S}_u} s_{j,k},
\end{equation}
where $\mathcal{S}_u$ contains the scored actions in $\mathcal{G}_u$ and $\lambda = 0.5$ controls the weight of the scores. Actions scored above the group mean receive a larger advantage, and actions scored below it receive a smaller one. We do not divide the centered scores by their standard deviation, so that a group in which the Teacher assigns similar scores receives only small adjustments. Each action's advantage is assigned to all of its generated tokens. Appendix~\ref{app:training} gives the full training procedure.

\paragraph{Setup.}
We use JEV~(\inlinemodellogo{jev}) as the Teacher and train Qwen2.5-1.5B-Instruct and Qwen2.5-7B-Instruct~\citep{qwen2025qwen25technicalreport} as policies on ALFWorld and WebShop. We compare with GRPO trained on outcome rewards only, using the results reported by \citet{NEURIPS2025_420c9f77}. Each trained policy is evaluated with three inference seeds. Appendix~\ref{app:setup} specifies the task splits, action budgets, and evaluation seeds.

\begingroup
\setlength{\intextsep}{5pt}
\begin{table}[H]
\centering
\caption{Policy success rates (\%) on ALFWorld and WebShop, averaged over three evaluation seeds. GRPO results are from GiGPO~\citep{NEURIPS2025_420c9f77}. Bold marks the best result within each policy size.}
\label{tab:policy-results}
\begingroup
\small
\setlength{\tabcolsep}{1.1pt}
\renewcommand{\arraystretch}{1.08}
\newcommand{\meanscore}[2]{\makebox[1.8em][r]{#1}\makebox[1.9em][l]{\ensuremath{_{\scriptscriptstyle\pm #2}}}}
\newcommand{\bestscore}[2]{\makebox[1.8em][r]{\textbf{#1}}\makebox[1.9em][l]{\ensuremath{_{\scriptscriptstyle\pm #2}}}}
\resizebox{\linewidth}{!}{%
\begin{tabular}{@{}lccccccc|cc@{}}
\toprule
	& \multicolumn{7}{c|}{\textbf{ALFWorld}} & \multicolumn{2}{c}{\textbf{WebShop}} \\
\cmidrule(lr){2-8}\cmidrule(l){9-10}
	& \multicolumn{3}{c}{No dependency} & \multicolumn{3}{c}{With dependency} & & & \\
\cmidrule(lr){2-4}\cmidrule(lr){5-7}
Method & Pick & Look & Pick2 & Clean & Heat & Cool & All & Score & Succ. \\
\midrule
\multicolumn{10}{@{}l}{\textit{Qwen2.5-1.5B-Instruct}} \\
GRPO & \meanscore{85.3}{1.5} & \meanscore{53.7}{8.0} & \meanscore{53.5}{5.6} & \meanscore{84.5}{6.8} & \meanscore{78.2}{7.9} & \meanscore{59.7}{5.0} & \meanscore{72.8}{3.6} & \meanscore{75.8}{3.5} & \meanscore{56.8}{3.8} \\
Prefix + Value & \bestscore{95.2}{4.4} & \meanscore{66.7}{11.8} & \bestscore{73.6}{4.8} & \meanscore{81.5}{9.8} & \meanscore{77.1}{7.2} & \meanscore{66.7}{2.3} & \meanscore{79.0}{3.6} & \meanscore{77.2}{0.3} & \meanscore{67.7}{0.5} \\
Prefix + Progress & \meanscore{87.6}{1.6} & \meanscore{56.4}{4.4} & \meanscore{72.2}{4.8} & \meanscore{81.5}{3.7} & \meanscore{83.3}{7.2} & \meanscore{76.0}{4.0} & \meanscore{78.3}{0.4} & \meanscore{78.6}{0.4} & \meanscore{66.6}{0.2} \\
Hindsight + Value & \meanscore{84.8}{1.6} & \meanscore{59.0}{4.4} & \meanscore{48.6}{8.7} & \meanscore{72.8}{2.1} & \meanscore{68.8}{6.2} & \meanscore{69.3}{2.3} & \meanscore{69.3}{0.7} & \meanscore{82.2}{0.7} & \meanscore{68.3}{1.5} \\
\rowcolor{black!7}
\textbf{Hindsight + Progress} & \meanscore{90.5}{1.6} & \bestscore{74.4}{11.8} & \meanscore{65.3}{6.4} & \bestscore{88.9}{7.4} & \bestscore{89.6}{3.6} & \bestscore{82.7}{2.3} & \bestscore{82.9}{2.9} & \bestscore{84.3}{0.4} & \bestscore{71.4}{0.7} \\
\midrule
\multicolumn{10}{@{}l}{\textit{Qwen2.5-7B-Instruct}} \\
GRPO & \meanscore{90.8}{5.1} & \meanscore{66.1}{6.7} & \meanscore{64.7}{7.3} & \meanscore{89.3}{5.4} & \meanscore{74.7}{6.9} & \meanscore{72.5}{5.4} & \meanscore{77.6}{5.2} & \meanscore{79.3}{2.8} & \meanscore{66.1}{3.7} \\
\rowcolor{black!7}
\textbf{Hindsight + Progress} & \bestscore{91.4}{5.7} & \bestscore{74.4}{11.8} & \bestscore{75.0}{7.2} & \bestscore{92.6}{0.0} & \bestscore{77.1}{9.5} & \bestscore{81.3}{2.3} & \bestscore{83.8}{1.6} & \bestscore{82.2}{0.7} & \bestscore{73.6}{0.3} \\
\bottomrule
\end{tabular}%
}
\endgroup
\end{table}
\endgroup

\subsubsection{Which Scores Make a Good Teacher?}
\label{sec:teacher-scores}

We train the 1.5B policy with all four combinations of question and context (Table~\ref{tab:policy-results}). Hindsight + Progress achieves the highest success on both environments, raising success over GRPO from 72.8\% to 82.9\% on ALFWorld and from 56.8\% to 71.4\% on WebShop. It improves over GRPO on every ALFWorld task type, with the largest gains on \textit{Cool} (+23.0\%) and \textit{Look} (+20.7\%). The improvement also holds for the 7B policy, raising success from 77.6\% to 83.8\% on ALFWorld and from 66.1\% to 73.6\% on WebShop. The 1.5B policy trained with the Teacher outperforms the 7B policy trained with GRPO on both environments.

These results are consistent with the Judge analyses. Progress helps distinguish the contributions of different actions (Section~\ref{sec:judge-question}), and Hindsight helps JEV recognize how actions support later progress toward the goal (Section~\ref{sec:judge-context}). Under Hindsight, Value assigns similar scores within a trajectory, scoring milestone actions only 0.05 higher than no-op actions (Figure~\ref{fig:jev-evaluator-objective}c). Its scores thus reveal little about which actions helped.

\begin{table}[H]
\vspace{-4pt}
\centering
\setlength{\belowcaptionskip}{2pt}
\caption{Task completion by action cutoff (\%) for 1.5B policies.}
\label{tab:completion-steps}
\begingroup
\small
\setlength{\tabcolsep}{3pt}
\begin{tabularx}{\linewidth}{@{}l*{7}{>{\raggedleft\arraybackslash}X}@{}}
\toprule
	& \multicolumn{4}{c}{ALFWorld} & \multicolumn{3}{c}{WebShop} \\
\cmidrule(lr){2-5}\cmidrule(l){6-8}
Method & $\leq 10$ & $\leq 15$ & $\leq 20$ & $\leq 30$ & $\leq 5$ & $\leq 10$ & $\leq 15$ \\
\midrule
Prefix + Value & $42.4_{\pm 1.1}$ & $59.3_{\pm 3.7}$ & $69.5_{\pm 5.1}$ & $79.0_{\pm 3.6}$ & $33.7_{\pm 1.1}$ & $60.9_{\pm 0.7}$ & $67.7_{\pm 0.5}$ \\
Prefix + Progress & $40.7_{\pm 0.7}$ & $56.0_{\pm 0.4}$ & $65.5_{\pm 0.4}$ & $78.3_{\pm 0.4}$ & $45.4_{\pm 1.4}$ & $61.9_{\pm 0.8}$ & $66.6_{\pm 0.2}$ \\
Hindsight + Value & $41.0_{\pm 3.4}$ & $58.6_{\pm 1.9}$ & $63.6_{\pm 0.7}$ & $69.3_{\pm 0.7}$ & $27.7_{\pm 2.1}$ & $62.2_{\pm 2.4}$ & $68.3_{\pm 1.5}$ \\
\rowcolor{black!7}
Hindsight + Progress & $47.4_{\pm 2.2}$ & $65.7_{\pm 1.2}$ & $74.5_{\pm 2.2}$ & $82.9_{\pm 2.9}$ & $50.7_{\pm 1.1}$ & $71.1_{\pm 0.4}$ & $71.4_{\pm 0.7}$ \\
\bottomrule
\end{tabularx}
\endgroup
\vspace{-7pt}
\end{table}

The advantage of Hindsight + Progress is largest on tasks with dependency, consistent with our analyses in Sections~\ref{sec:jev-actor} and~\ref{sec:judge-context}. On these tasks, it reaches 86.8\% and exceeds the other settings by 6.9\% to 16.2\%. Hindsight + Progress also completes tasks in fewer actions (Table~\ref{tab:completion-steps}). Within 20 actions, it completes 74.5\% of ALFWorld tasks, compared with 63.6\% to 69.5\% for the other settings. On WebShop, it completes 71.1\% of tasks within 10 actions, close to its final success rate, compared with 60.9\% to 62.2\%.

\Needspace{21\baselineskip}
\vspace{4pt}
\paragraph{Teacher API costs.}
We query JEV~(\inlinemodellogo{jev}) through OpenRouter to score each executed action during training. Table~\ref{tab:teacher-api-costs} reports the training settings, JEV token prices~\citep{openrouterJev113Pricing}, and Teacher API costs for the four Hindsight + Progress policies.

\begin{table}[H]
\vspace{-6pt}
\centering
\setlength{\belowcaptionskip}{2pt}
\caption{Training settings and Teacher API costs for the four Hindsight + Progress policies.}
\label{tab:teacher-api-costs}
\begingroup
\small
\setlength{\tabcolsep}{4pt}
\renewcommand{\arraystretch}{1.0}
\begin{tabularx}{\linewidth}{@{}l*{4}{>{\centering\arraybackslash}X}@{}}
\toprule
	& \multicolumn{2}{c}{\textbf{ALFWorld}} & \multicolumn{2}{c}{\textbf{WebShop}} \\
\cmidrule(lr){2-3}\cmidrule(l){4-5}
Training setting & 1.5B & 7B & 1.5B & 7B \\
\midrule
Training steps & 150 & 150 & 150 & 150 \\
Tasks per batch & 16 & 16 & 16 & 16 \\
Rollouts per task & 8 & 8 & 8 & 8 \\
Max actions per rollout & 30 & 30 & 15 & 15 \\
Max policy prompt / response tokens & 2,048 / 512 & 2,048 / 512 & 4,096 / 512 & 4,096 / 512 \\
\begin{tabular}[c]{@{}l@{}}JEV 1.13 input / output price\\(USD per million tokens)\end{tabular} & \$0.042 / \$0 & \$0.042 / \$0 & \$0.042 / \$0 & \$0.042 / \$0 \\
\midrule
\rowcolor{black!7}
\textbf{Teacher API cost (USD)} & \textbf{30.07} & \textbf{24.50} & \textbf{22.21} & \textbf{26.83} \\
\bottomrule
\end{tabularx}
\endgroup
\vspace{-6pt}
\end{table}

\vspace{5pt}
\subsubsection{Does a Good Teacher Need to Act Well?}
\label{sec:teacher-actor}

On the same ALFWorld evaluation tasks, JEV~(\inlinemodellogo{jev}) completes 42.9\% as an Actor, whereas the 1.5B policy trained with its scores completes 82.9\%, including tasks with dependencies, on which JEV performs worst as an Actor (Figure~\ref{fig:actor-task-heatmap}). The Teacher thus does not need to solve a task itself to provide a useful training signal for it.

\vspace{4pt}
This gap reflects the difference between the two roles. On tasks with dependent subgoals, an easier subgoal draws Actors away from the actions needed to complete the subgoal that ought to be met first (Section~\ref{sec:jev-actor}). Yet these actions can receive greater credit when the same models judge their contribution after seeing the whole trajectory: the later steps reveal how they support the subgoal that needs to be met first (Section~\ref{sec:judge}). We use these judgments as Teacher scores, giving the policy additional supervision about which actions contribute to the goal. The resulting improvement on tasks with dependency connects the three roles: a decision model that struggles to choose the required actions can still recognize their contribution when observing full trajectories and provide useful supervision for learning to act.

\vspace{6pt}
\section{Takeaways for Using Decision Models in Agents}
\label{sec:recommendations}

Beyond the above findings, our analyses suggest several lessons for using decision models in agents.

\begin{itemize}[leftmargin=*,topsep=3pt,itemsep=2pt,parsep=0pt]
		\item \textbf{Actors know the required step but are drawn to the easier one.} Actors choose the step that must come first when it is the only subgoal in the goal, and general reminders do not prevent the error when an easier subgoal is present. When a decision model must act, restructuring the goal, for example into ordered subgoals, is more promising than merely adding instructions to the prompt (Section~\ref{sec:jev-actor}).
		\item \textbf{The question sets the granularity of the signal.} A question about the eventual outcome gives similar scores to all actions in a trajectory and largely repeats the outcome reward, while a question about each action's contribution separates actions within a trajectory (Sections~\ref{sec:judge-question} and~\ref{sec:teacher-scores}).
		\item \textbf{Hindsight judgments target the actor's own failure.} Seeing the whole trajectory helps Judges recognize the contributions of actions toward the subgoal that needs to be met first, which Actors may overlook in favor of an easier subgoal. Decision models are therefore best used to score actions after an episode ends, as in RL training, and are less suited to filtering actions online, where only the prefix is available (Sections~\ref{sec:judge-context} and~\ref{sec:teacher-actor}).
		\item \textbf{Judges can be screened before training.} The Judge measures on recorded trajectories identified the question and context that trained the best policy, at a small fraction of the cost of an RL run. Compare candidate judges and questions this way first, and choose by judging quality, not acting ability (Sections~\ref{sec:judge} and~\ref{sec:teacher-actor}).
		\item \textbf{Step-level supervision becomes affordable.} Decision models judge more accurately and at lower cost than prompted general LLMs, and the Teacher is not needed at test time. Scoring every action during RL training is therefore practical, and can let a small policy outperform a larger one trained with outcome rewards alone (Section~\ref{sec:teacher-scores} and Appendix~\ref{sec:judge-cost}).
\end{itemize}

\vspace{6pt}
\section{Related Work}
\label{sec:related-work}

\paragraph{Decision models.}
JEV~\citep{typesafe2026jevlaunch} takes a question and a context and returns a choice among options or a score, providing a single interface for both selecting and evaluating actions. Similar models include GPT Decisions~\citep{openai2026decisions} and Decider v1.2~\citep{perplexity2026deciderv12}. Prior work mainly uses decision models to act. \citet{yang2026codeownssimulationjev} study JEV's action selection, finding failures caused by surface shortcuts and improvements from explicit subgoals and simulated action outcomes. JevAgentHarness~\citep{zhu2026jevagentharnessbenchmarkingdeployingtraining} trains decision models with correct-option labels and JEV's option probabilities, and lets them share online control with an LLM. We instead study decision models in three roles, focusing as Actors on goals whose subgoals must be met in order, and show that their performance as Actors does not reflect their value as Judges and Teachers.

\vspace{6pt}
\paragraph{LLM judges and process supervision.}
LLM judges~\citep{NEURIPS2023_91f18a12,ICLR2026_e812af67} and AI feedback~\citep{bai2022constitutionalaiharmlessnessai,zhang-etal-2024-self-contrast} provide evaluation without task-specific training, but they are mostly used to compare final responses, not to score every step of a long trajectory. Process reward models score intermediate steps and improve over outcome supervision in mathematical reasoning~\citep{ICLR2024_aca97732,he-etal-2024-advancing}, but they require step-level labels for each domain. Our Judge analysis shows that decision models can score individual actions without such labels, and that the question asked and the context shown determine how useful the scores are.

\vspace{6pt}
\paragraph{Credit assignment for LLM agents.}
LLM agents act through long interactions with their environment~\citep{yao2023react,shridhar2021alfworld,ICLR2024_4410c071} and are commonly trained with outcome rewards using group-based policy optimization such as GRPO~\citep{shao2024deepseekmathpushinglimitsmathematical,wang-etal-2026-agent-g2}, where every step of a trajectory receives the same advantage. Recent work derives finer credit from the structure of the rollouts~\citep{NEURIPS2025_420c9f77,pmlr-v306-wang26jm}; GiGPO, for example, compares actions taken from the same state across trajectories. Hindsight credit assignment~\citep{NEURIPS2019_195f1538} uses what happens after an action to judge its contribution. We obtain step-level credit from a frozen decision model and find it most useful when the model, as in hindsight credit assignment, sees the rest of the trajectory. Because the Teacher does not need to solve the task itself, our setting also relates to weak-to-strong generalization~\citep{pmlr-v235-burns24b}, where a strong model trained on labels from a weaker one can surpass it.

\par
\vspace{10pt}
\section{Conclusion}
\label{sec:conclusion}

We presented a systematic study of decision models for long-horizon agents in three roles: Actor, Judge, and Teacher. As Actors, decision models often satisfy an easier subgoal before the one that needs to be met first, and a general reminder in the prompt does not reliably prevent this. As Judges, the same models can recognize the contributions of actions they overlook when acting, provided they are asked whether each action helped achieve the goal and shown the whole trajectory. As Teachers, judgments made this way improve policy training on both ALFWorld and WebShop, including on tasks that the Teacher rarely completes as an Actor. Together, these results show that a decision model's ability to act does not fully reflect its value for agent training, which depends on the question it is asked and the context it is given. These judgments provide step-level supervision during training without requiring additional Teacher calls at test time.

\section{Limitations and Future Work}
\label{sec:limitations}

Building on our findings, we identify several directions for future work:
\begin{itemize}[leftmargin=*]
		\item \textbf{Longer and more varied tasks.} All experiments use two text-based environments with episodes of at most 30 actions. Software engineering, tool use, and assistant tasks involve longer trajectories and more varied dependencies. Whether asking about each action's contribution with the full trajectory visible remains effective in these settings is an important open question.
		\item \textbf{Online judging.} Our judges score actions after an episode ends. When a judge is used online, for example to filter or rerank actions, only the prefix is available. Methods that delay or revise scores as later steps arrive could extend hindsight credit to such settings.
		\item \textbf{Training decision models as judges.} JEV and GPT Decisions are closed models with undisclosed adaptation procedures, so we cannot tell which training choices make a decision model a good judge. Open-weight decision models such as Decider v1.2 make it possible to train directly for judging the contribution of individual actions, the setting in which decision models are most useful as teachers.
\end{itemize}

\bibliographystyle{plainnat}
\bibliography{references}

\appendix
\clearpage
\section{Benchmark Setup}
\label{app:setup}

We adopt the training and evaluation splits from GiGPO~\citep{NEURIPS2025_420c9f77} for both ALFWorld and WebShop. For both benchmarks, we evaluate each trained policy on the same tasks using three different inference seeds (0, 1, and 2). We use a budget of 30 actions per evaluation episode on ALFWorld and 15 on WebShop.

\begin{table}[H]
\centering\footnotesize
\setlength{\tabcolsep}{4pt}
\caption{Model identifiers, interfaces, and API providers for the six models evaluated in this work.}
\label{tab:decision-model-versions}
\begin{tabular*}{\textwidth}{@{\extracolsep{\fill}}llll@{}}
\toprule
Model & Model identifier & Interface or readout & API provider \\
\midrule
\modellogo{jev}\, JEV & \texttt{typesafe/jev-1.13-20260917} & Native decision interface & OpenRouter \\
\modellogo{decider}\, Decider v1.2 & \texttt{perplexity/pplx-decider-v1.2-20261016} & Native decision interface & OpenRouter \\
\modellogo{gpt-decisions}\, GPT Decisions & \texttt{openai/gpt-6-luna-decisions-20261006} & Native decision interface & OpenRouter \\
\modellogo{qwen-prompted}\, Qwen prompted & \texttt{qwen/qwen3.8-27b} & Prompting & OpenRouter \\
\modellogo{qwen-constrained}\, Qwen constrained & \texttt{qwen/qwen3.8-27b} & Constrained readout & OpenRouter \\
\modellogo{gpt-prompted}\, GPT Luna & \texttt{openai/gpt-6-luna} & Prompting & OpenRouter \\
\bottomrule
\end{tabular*}
\end{table}

\section{Decision-Model Interfaces and Prompts}
\label{app:prompts}

Each request separates the \textbf{user input} from the \textbf{system prompt}. The user input contains the \textbf{Context}, which specifies the visible information, and the \textbf{Question}, which specifies what to assess. Actor requests also contain the available actions. Dedicated decision models receive the context, question instructions, and answer specification through structured API fields, with no system prompt supplied. Prompted and constrained LLMs receive the same content in a user message with separate Context and Question sections; their system prompts specify how to return the answer.

\begingroup
\tcbset{breakable=false}
\begin{promptbox}[Request Structure]
\begin{verbatim}
User input
  Context: task and visible observations/actions
  Question: instructions
  Available options: candidate actions (Actor only)

System prompt
  Dedicated Decision Models: none supplied
  Prompted: JSON response instructions
  Constrained: single-token response instructions
\end{verbatim}
\end{promptbox}

For prompted and constrained LLMs, the user message takes the following form. The question key is \texttt{next\_action}, \texttt{value}, or \texttt{progress}. The available-options section is included only for Actor requests. Constrained requests additionally end with \texttt{Answer:}.

\begin{promptbox}[User Message Structure]
\begin{verbatim}
Question key: <question_key>

Context: <context JSON>

Question: <question instructions>

Available options: <candidate options JSON>
\end{verbatim}
\end{promptbox}

Table~\ref{tab:decision-model-versions} lists the six models used in the Actor and Judge experiments. Reasoning is disabled for Qwen prompted~(\inlinemodellogo{qwen-prompted}), Qwen constrained~(\inlinemodellogo{qwen-constrained}), and GPT Luna~(\inlinemodellogo{gpt-prompted}).

\subsection{Actor Interface and Prompts}
\label{app:actor-interfaces}

\subsubsection{User Prompt}
The Actor selects the next action from valid commands. Its context contains the task, past observation--action pairs, current observation, and action budget. The current observation precedes the action to be selected. Candidate actions are supplied separately from the context.

\begin{promptbox}[Actor Context][top=7pt,bottom=5pt,before skip=10pt,after skip=6pt]
\begin{verbatim}
{
  "task": "<task goal>",
  "history": [
    {"observation": "<past observation>", "action": "<past action>"}, ...],
  "current_observation": "<current observation>",
  "steps_taken": <number of executed actions>,
  "steps_remaining": <remaining budget, including the next action>
}
\end{verbatim}
\end{promptbox}
\begin{promptbox}[Actor Question][top=7pt,bottom=5pt,before skip=10pt,after skip=6pt]
Which available action should be executed next to maximize the chance of completing the task within the remaining action budget, including this action? Use only the task, observed history, current observation, and the available actions.
\end{promptbox}
\begin{promptbox}[Available Options][top=7pt,bottom=5pt,before skip=10pt,after skip=6pt]
\begin{verbatim}
[
  {"id": "a0", "action": "<candidate action 0>"},
  {"id": "a1", "action": "<candidate action 1>"},
  ...
]
\end{verbatim}
\end{promptbox}

\subsubsection{System Prompt}
\paragraph{Dedicated Decision Models.}
None supplied.

\Needspace{24\baselineskip}
\paragraph{Prompted.}
The following system prompt requests the choice and its probabilities in JSON.

\begin{promptbox}[Prompted Actor System Prompt][top=7pt,bottom=5pt,before skip=10pt,after skip=6pt]
You select the next action for an agent.

Answer the question using only the supplied context and available options.

Assign a probability to every option, expressing your estimate that it is the best next action under the supplied question. These probabilities describe the relative preference among the options, not the probability of eventual task success.

Return a single JSON object with this structure:
\begin{verbatim}
{"answers": {
    "<question_key>": {
      "type": "choice",
      "choice": "<selected_option_id>",
      "probabilities": {"<option_id>": 0.0},
      "confidence": 0.0
    }
}}
\end{verbatim}

Use the supplied question key and option IDs exactly. Include every option ID exactly once in probabilities. Each probability must be between 0 and 1, and their sum must equal 1. Select the option with the highest probability. Break ties by choosing the option appearing first in the supplied options. Set confidence to the selected option's probability.

Do not include additional fields, explanations, or Markdown fences.
\end{promptbox}

\paragraph{Constrained.}
We map candidate IDs to single-token labels such as \texttt{A}, \texttt{B}, and \texttt{C}, preserving the corresponding commands. These labels appear in the available-options section. We read the probabilities at the first answer position and select the highest-probability returned option label.

\begin{promptbox}[Constrained Actor System Prompt]
Answer the supplied question using only the supplied context and available options. Each option ID is an opaque, case-sensitive label. Return exactly the ID of the best option as a single token. Do not include spaces, explanations, punctuation, JSON, or Markdown.
\end{promptbox}

In all three cases, we execute the chosen option and normalize the option probabilities within each response for the action-probability analysis (Section~\ref{sec:actor-goal-conditions}).

\subsubsection{Actor Analysis}
The following variants correspond to the goal-masking and dependency-hint analyses in Sections~\ref{sec:actor-goal-conditions} and~\ref{sec:actor-condition-dependency}.

\paragraph{Goal masking.}
In the goal-masking experiment (Section~\ref{sec:actor-goal-conditions}), we use the standard question with either the full goal or the goal with the placement subgoal removed. For example, ``put a heated mug on the side table'' becomes ``heat a mug''.

\paragraph{Dependency hint and ordering example.}
In the dependency-hint experiment (Section~\ref{sec:actor-condition-dependency}), all three conditions retain the full goal and use their assigned prompt at every decision. \textit{Original} uses the Actor Question. \textit{Hint} appends the following sentence: ``Consider possible dependencies among task requirements''. \textit{Hint + example} appends the following passage after this hint.

\begin{promptbox}[Dependency Example]
For example, the task is to place a heated mug on a table. Initially, the mug has not been heated and is not on the table. Heating the mug and placing it on the table are two task requirements, and they have a dependency: to finish the task, you need to first heat the mug, then place the heated mug on the table.
\end{promptbox}

\subsection{Judge Interface and Prompts}
\label{app:evaluator-interfaces}

\subsubsection{User Prompt}
The Judge scores one executed action in $[0,1]$. We construct its user input by selecting a context and a question independently.

\paragraph{Context.}
Both contexts use the same \texttt{goal} and \texttt{trajectory} fields. Each trajectory entry pairs an observation with the action taken after it. \textit{Prefix} ends with the target action and excludes the observation returned after its execution. \textit{Hindsight} includes the whole trajectory and a final observation-only entry with \texttt{action=null}. For a target action at index $t$, the visible sequences are
\[
\begin{aligned}
\text{Prefix:}\quad &(o_0,a_0),\ldots,(o_t,a_t),\\
\text{Hindsight:}\quad &(o_0,a_0),\ldots,(o_t,a_t),\ldots,(o_T,\texttt{null}).
\end{aligned}
\]
The context is serialized as follows, with entries retained according to the selected visibility range.

\begin{promptbox}[Judge Context]
\begin{verbatim}
{
  "goal": "<task goal>",
  "trajectory": [
    {"obs": "<observation before action 0>", "action": "<action 0>"},
    ...
  ]
}
\end{verbatim}
\end{promptbox}

The history contains environment observations and executed actions, without the agent's internal reasoning. Environment-provided success labels are masked. In ALFWorld, the final observation is retained because it describes the last action's outcome without explicitly reporting task success. For WebShop, the terminal purchase-page observation is replaced with ``Shopping process has ended.'' to remove the environment-provided score.

\paragraph{Question.}
Both Value and Progress begin with the same target-action instructions. The target index refers to the visible trajectory array and starts at zero.

\begin{promptbox}[Target-Action Instructions]
Evaluate only the action at trajectory[\{target\_step\}]. Trajectory indices start at 0. Each obs is observed before its paired action; the next obs contains feedback after that action. An entry with action=null contains observation only.
\end{promptbox}

For \textbf{Value}, we append the remaining action budget and the following question. The budget counts the target action itself.

\begin{promptbox}[Value Question]
Steps remaining before max\_steps: \{remaining\_steps\} (including the current action).

Question: What is the probability that the task will be completed within the remaining step budget if this action is executed first and the same agent policy then continues? Estimate the success probability conditioned on this action, not the probability of choosing the action. The remaining step budget includes the current action.
\end{promptbox}

\Needspace{9\baselineskip}
For \textbf{Progress}, we append the following question.

\begin{promptbox}[Progress Question]
Question: Did the target action make a useful contribution toward completing the goal?

Later failure does not erase earlier useful progress. Later success does not make an unnecessary detour useful.
\end{promptbox}

For a given target action, the question text is the same under Prefix and Hindsight; only the visible trajectory changes.

\subsubsection{System Prompt}
\paragraph{Dedicated Decision Models.}
None supplied.

\Needspace{19\baselineskip}
\paragraph{Prompted.}
The following system prompt requests a JSON score, which we use directly.

\begin{promptbox}[Prompted Judge System Prompt]
Answer the supplied question with a probability.

Use only the supplied context, question and answer instructions.

Return a single JSON object with this structure:
\begin{verbatim}
{"answers": {
    "<question_key>": {
      "type": "score",
      "value": 0.0
    }
}}
\end{verbatim}

Use the supplied question key exactly. Set value to a finite JSON number between 0 and 1, inclusive. Include only the supplied question key in answers. Do not include additional fields, explanations, or Markdown fences.
\end{promptbox}

\Needspace{10\baselineskip}
\paragraph{Constrained.}
The following system prompt requests a single Yes/No token.

\begin{promptbox}[Constrained Judge System Prompt]
Answer the supplied question using only the supplied context. If asked for a probability, judge whether the described event occurs. Return exactly one token: Yes or No. Do not include spaces, explanations, punctuation, JSON, or Markdown.
\end{promptbox}

We normalize the first-position token probabilities over \texttt{Yes} and \texttt{No} and use the resulting probability of \texttt{Yes} as the score:
\[
 s = \frac{P(\texttt{Yes})}{P(\texttt{Yes})+P(\texttt{No})}.
\]

\subsection{Teacher Interface}

We use JEV's~(\inlinemodellogo{jev}) native scoring interface on rollouts from the policy being trained. The Teacher uses the context formats and question templates in Appendix~\ref{app:evaluator-interfaces}, with one request per executed action. Appendix~\ref{app:training} describes how the scores enter the policy update.

\endgroup

\section{Analysis Protocols}

\subsection{Action-Type Annotation}
\label{app:evaluator-protocols}
We identify milestone and no-op actions from recorded environment states and action transitions. A \textbf{milestone} action achieves a subgoal, a definition we adopt directly from \citet{pmlr-v306-wang26jm}. \textbf{No-op} actions require an unchanged environment state together with repeated observation--action behavior or feedback indicating an ineffective command. An unchanged environment state indicates that the action has not altered the environment. Repeated observations indicate that the action has not provided new information about the environment. In the question comparison, actions outside these two categories are grouped as Other.

\subsection{Judge scoring costs}
\label{app:evaluator-efficiency-protocol}
For a fair comparison of scoring costs, we query all six Judges through the OpenRouter API. For each Judge, we average the charges reported by OpenRouter for the successful requests that produced the action scores and multiply the result by 1,000 to obtain the cost per 1,000 action scores.

\clearpage
\section{Decision-Model Supervision: Training Implementation}
\label{app:training}

Algorithm~\ref{alg:teacher-training} summarizes policy training with a fixed Teacher. Table~\ref{tab:training-hyperparameters} lists the training hyperparameters; the Teacher inputs and prompts are given in Appendix~\ref{app:prompts}.

We use veRL's PPO-style policy update~\citep{10.1145/3689031.3696075,schulman2017proximalpolicyoptimizationalgorithms} with dual clipping, a reference-policy KL penalty, and an entropy bonus, denoted by $J_{\mathrm{PPO}}$.

\begin{algorithm}[H]
\caption{Policy Training with a Fixed Decision-Model Teacher}
\label{alg:teacher-training}
\small
\begin{algorithmic}[1]
\Require Policy $\pi_\theta$, fixed Teacher $D$, reference policy $\pi_{\mathrm{ref}}$, group size $G$, Teacher weight $\lambda$
\For{each training iteration}
	\State $\pi_{\mathrm{old}} \gets \pi_\theta$; sample a batch of tasks
	\For{each task $u$ in the batch}
		\State {\color[rgb]{0,0.6,0}// Collect trajectories and outcome advantages}
		\State Sample $\mathcal G_u=\{\tau_i\}_{i=1}^{G}$ using $\pi_{\mathrm{old}}$, where trajectory $\tau_i$ contains $T_i$ actions
		\State Record $R_{i,t}$, the outcome return for action $t$ of trajectory $i$, for $i=1,\ldots,G$ and $t=1,\ldots,T_i$
		\State $\mathcal T_u \gets \{(i,t):1\le i\le G,\ 1\le t\le T_i\}$, the set of executed-action indices
		\State $\mu_u \gets \operatorname{mean}_{\mathcal T_u} R_{i,t}$; $\sigma_u \gets \operatorname{sample\ std}_{\mathcal T_u} R_{i,t}$
		\State $A^{\mathrm{out}}_{i,t} \gets (R_{i,t}-\mu_u)/(\sigma_u+10^{-6})$ for all $(i,t)\in\mathcal T_u$
		\State {\color[rgb]{0,0.6,0}// Score actions after rollout and center Teacher scores}
		\State Obtain Teacher score $s_{i,t}$ for each $(i,t)\in\mathcal T_u$ using the selected question and context
		\State $\bar s_u\gets |\mathcal T_u|^{-1}\sum_{(i,t)\in\mathcal T_u}s_{i,t}$
		\State {\color[rgb]{0,0.6,0}// Combine advantages}
		\State $A_{i,t}\gets A^{\mathrm{out}}_{i,t}+\lambda(s_{i,t}-\bar s_u)$ for all $(i,t)\in\mathcal T_u$
	\EndFor
	\State {\color[rgb]{0,0.6,0}// Policy update}
	\State Assign $A_{i,t}$ to each token of action $t$ in trajectory $i$
	\State Update $\theta$ by maximizing $J_{\mathrm{PPO}}(\theta)$ using the combined advantages $A_{i,t}$
\EndFor
\end{algorithmic}
\end{algorithm}

\begin{table}[H]
\centering\footnotesize
\caption{Training hyperparameters for the policies. Paired values denote ALFWorld / WebShop. Other settings follow GiGPO~\citep{NEURIPS2025_420c9f77}.}
\label{tab:training-hyperparameters}
\setlength{\tabcolsep}{3pt}
\renewcommand{\arraystretch}{1.04}
\begin{tabularx}{\textwidth}{@{}Xc@{\hspace{14pt}}Xc@{}}
\toprule
\textbf{Hyperparameter} & \textbf{Value} & \textbf{Hyperparameter} & \textbf{Value} \\
\midrule
\rowcolor{black!7}
\multicolumn{2}{l}{\textbf{Teacher supervision}} & \multicolumn{2}{l}{\textbf{Batch configuration}} \\
Teacher advantage weight $\lambda$ & $0.5$ & Tasks per iteration & $16$ \\
Teacher-score normalization & Mean centering only & Rollouts per task $G$ & $8$ \\
Default question / context & Progress / Hindsight & PPO mini-batch size & $256$ / $64$ \\
& & Training iterations & $150$ \\
\midrule
\rowcolor{black!7}
\multicolumn{2}{l}{\textbf{Optimization}} & \multicolumn{2}{l}{\textbf{Sequence and environment}} \\
Learning rate & $1\times10^{-6}$ & Max policy prompt tokens & $2{,}048$ / $4{,}096$ \\
PPO clip ratio $\varepsilon$ & $0.2$ & Max policy response tokens & $512$ \\
Dual-clip coefficient & $3.0$ & Policy history length (steps) & $2$ \\
KL loss coefficient $\beta$ & $0.01$ & Max actions per rollout & $30$ / $15$ \\
Entropy coefficient $\eta$ & $0.001$ & Invalid-action penalty & $0.1$ \\
Outcome normalization stabilizer $\epsilon$ & $10^{-6}$ & & \\
\bottomrule
\end{tabularx}
\end{table}

\Needspace{25\baselineskip}
\suppressfloats[t]
\section{Supplementary Results}
\label{app:additional-results}

\subsection{Actor Performance on Broader ALFWorld Tasks}
\label{app:actor-performance-l1}
We additionally evaluate all six Actors on the ALFWorld tasks from the L1 split used by \citet{ICLR2026_baa41b73}. Table~\ref{tab:actor-performance-l1} reports these results.

\begin{table}[H]
\centering
\caption{Actor success rates (\%) on ALFWorld, using the same evaluation tasks as the L1 split in \citet{ICLR2026_baa41b73}. Each Actor is evaluated once per task. Bold marks the best result in each column.}
\label{tab:actor-performance-l1}
\small
\begin{tabular*}{\textwidth}{@{\extracolsep{\fill}}lccccccc@{}}
\toprule
	& \multicolumn{3}{c}{No dependency} & \multicolumn{3}{c}{With dependency} & \\
\cmidrule(lr){2-4} \cmidrule(lr){5-7}
Actor & Pick & Look & Pick2 & Clean & Heat & Cool & All \\
\midrule
\multicolumn{8}{l}{\textit{Dedicated decision models}} \\
\modellogo{jev}\, JEV & 75.0 & 66.7 & 76.5 & 9.7 & 13.0 & 14.3 & 38.8 \\
\modellogo{decider}\, Decider v1.2 & \textbf{91.7} & 66.7 & 76.5 & \textbf{67.7} & 26.1 & \textbf{57.1} & \textbf{64.2} \\
\modellogo{gpt-decisions}\, GPT Decisions & 79.2 & \textbf{94.4} & 58.8 & 48.4 & 26.1 & 33.3 & 55.2 \\
\midrule
\multicolumn{8}{l}{\textit{General LLMs as decision models}} \\
\modellogo{qwen-prompted}\, Qwen prompted & 62.5 & 88.9 & 52.9 & 25.8 & 8.7 & 9.5 & 38.8 \\
\modellogo{qwen-constrained}\, Qwen constrained & \textbf{91.7} & 72.2 & \textbf{82.4} & 54.8 & \textbf{30.4} & 23.8 & 58.2 \\
\modellogo{gpt-prompted}\, GPT Luna & 87.5 & \textbf{94.4} & 76.5 & 38.7 & \textbf{30.4} & 19.0 & 55.2 \\
\bottomrule
\end{tabular*}
\end{table}

\subsection{Action-score heatmaps across ALFWorld Judges}
\label{app:judge-heatmaps}
\begin{figure}[H]
	\centering
	\includegraphics[width=\linewidth]{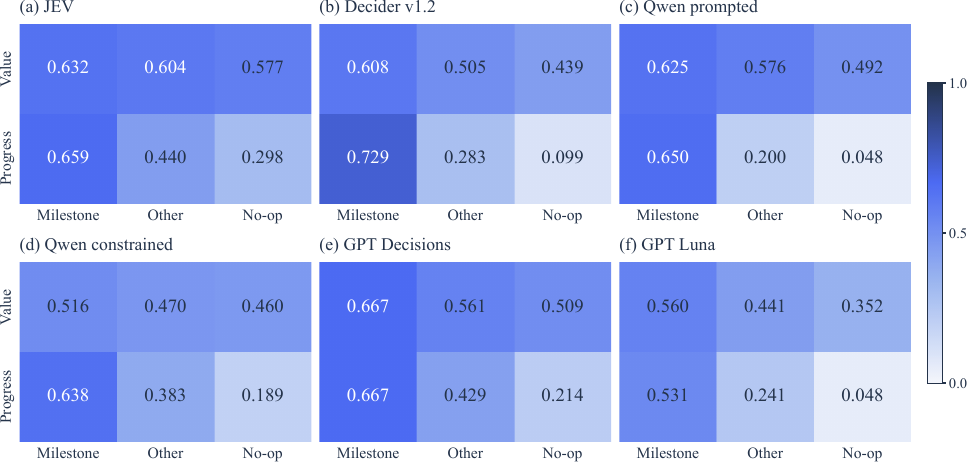}\par
	\caption{\textbf{Action scores across six Judges on ALFWorld.} Mean Hindsight scores under Value and Progress on the same 462 trajectories as Figure~\ref{fig:jev-evaluator-objective}. Scores are averaged within each trajectory and then across trajectories. All panels share the $[0,1]$ scale.}
	\label{fig:alfworld-evaluator-heatmaps}
\end{figure}

Figure~\ref{fig:alfworld-evaluator-heatmaps} compares all six Judges on the same 462 ALFWorld trajectories used for the JEV analysis in Section~\ref{sec:judge-question}. Each cell averages scores within an action group in each trajectory and then averages across trajectories with equal weight. For all six Judges, Progress yields a larger gap between milestone and no-op scores than Value.

\clearpage
\subsection{Action rankings across visible contexts}
\label{app:judge-context-models}
\begin{figure}[H]
\centering
\includegraphics[width=\linewidth]{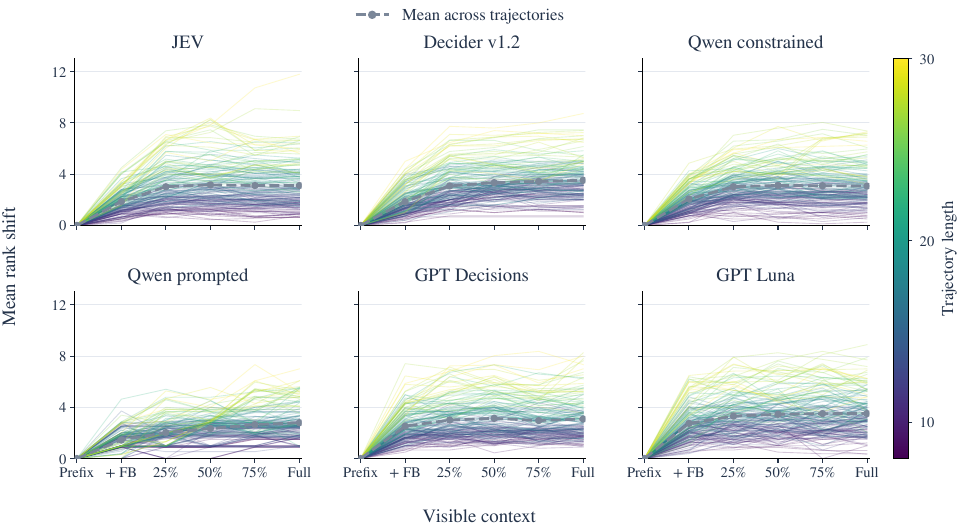}
\caption{\textbf{Action rankings as more context becomes visible across six Judges.} Each thin line shows a trajectory's mean rank shift relative to Prefix; the gray dashed line is the mean across trajectories. FB denotes immediate feedback.}
\label{fig:judge-context-models}
\end{figure}

Figure~\ref{fig:judge-context-models} extends the context-expansion analysis in Section~\ref{sec:judge-context} to all six Judges.

\subsection{Complete trajectory for the context-expansion example}
\label{app:context-example}
Table~\ref{tab:context-example} gives the complete action sequence and JEV scores for the heated-mug example in Figure~\ref{fig:judge-context-case}.

\begin{table}[H]
\centering\scriptsize
\caption{JEV's Progress scores for the complete heated-mug trajectory. Contexts follow Figure~\ref{fig:judge-context-case}.}
\label{tab:context-example}
\setlength{\tabcolsep}{2pt}
\begin{tabular*}{\textwidth}{@{\extracolsep{\fill}}rlrrrrrr@{}}
\toprule
Step & Action & Prefix & + FB & 25\% & 50\% & 75\% & Full \\
\midrule
1 & look & 0.47 & 0.57 & 0.62 & 0.61 & 0.56 & 0.54 \\
2 & look & 0.25 & 0.25 & 0.24 & 0.23 & 0.24 & 0.23 \\
3 & go to coffeemachine 1 & 0.49 & 0.72 & 0.46 & 0.40 & 0.46 & 0.44 \\
4 & take mug 3 from coffeemachine 1 & 0.92 & 0.92 & 0.90 & 0.82 & 0.83 & 0.85 \\
5 & go to cabinet 1 & 0.80 & 0.81 & 0.69 & 0.67 & 0.66 & 0.68 \\
6 & move mug 3 to cabinet 1 & 0.79 & 0.90 & 0.52 & 0.47 & 0.52 & 0.60 \\
7 & take mug 3 from cabinet 1 & 0.22 & 0.25 & 0.34 & 0.43 & 0.49 & 0.50 \\
8 & go to microwave 1 & 0.23 & 0.22 & 0.19 & 0.40 & 0.46 & 0.50 \\
9 & open microwave 1 & 0.20 & 0.18 & 0.66 & 0.66 & 0.76 & 0.76 \\
10 & heat mug 3 with microwave 1 & 0.42 & 0.67 & 0.47 & 0.47 & 0.70 & 0.70 \\
11 & go to cabinet 1 & 0.63 & 0.45 & 0.80 & 0.80 & 0.80 & 0.80 \\
12 & move mug 3 to cabinet 1 & 0.68 & 0.85 & 0.85 & 0.85 & 0.85 & 0.85 \\
\bottomrule
\end{tabular*}
\end{table}

\clearpage
\subsection{Cost and Ranking Quality of Action Judgments}
\label{sec:judge-cost}
\begin{figure}[htbp]
	\vspace{-4pt}
	\centering
	\setlength{\abovecaptionskip}{2pt}
	\includegraphics[width=\linewidth]{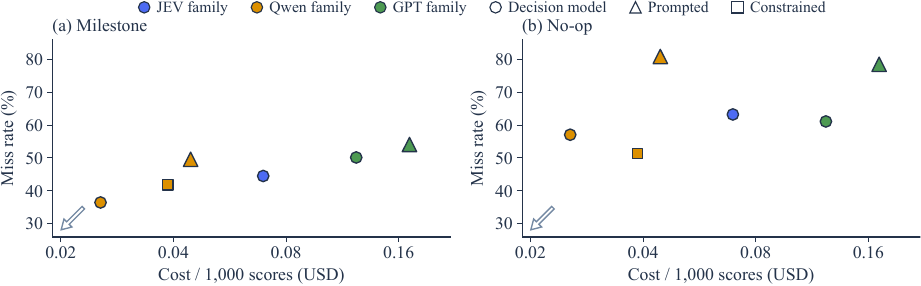}
	\caption{Miss rate versus cost for six Judges with Hindsight + Progress on 462 ALFWorld trajectories. (a) Milestone actions. (b) No-op actions. Cost is measured in USD per 1,000 action scores on a logarithmic scale. \textbf{Lower left is better}. Appendix~\ref{app:evaluator-efficiency-protocol} describes how scoring costs are calculated.}
	\label{fig:judge-cost}
	\label{fig:jev-evaluator-efficiency}
	\vspace{-4pt}
\end{figure}

Using a Judge as a Teacher requires scoring many actions during policy training. We therefore examine the cost and ranking quality of the Hindsight + Progress judgments used in Section~\ref{sec:teacher}. We compare all six Judges on the ALFWorld trajectories from Section~\ref{sec:judge-question}. For each trajectory with $R$ milestone actions, we measure the fraction of milestones missing from its $R$ highest-scored actions. Likewise, for each trajectory with $R$ no-op actions, we measure the fraction of no-ops missing from its $R$ lowest-scored actions. Figure~\ref{fig:judge-cost} plots these miss rates against the cost per 1,000 action scores.

Decider v1.2~(\inlinemodellogo{decider}) misses fewer milestone actions than Qwen prompted~(\inlinemodellogo{qwen-prompted}) (36.5\% vs.\ 49.6\%) and fewer no-op actions (57.1\% vs.\ 80.8\%), while costing about 43\% less per score. GPT Decisions~(\inlinemodellogo{gpt-decisions}) likewise misses fewer milestone actions than GPT Luna~(\inlinemodellogo{gpt-prompted}) (50.2\% vs.\ 54.1\%) and fewer no-op actions (61.1\% vs.\ 78.4\%), while costing about 28\% less per score. Although Qwen constrained~(\inlinemodellogo{qwen-constrained}) has the lowest no-op miss rate (51.3\%), it costs about 1.5 times as much per score as Decider v1.2 and has a higher milestone miss rate (41.8\% vs.\ 36.5\%). In both paired comparisons, the decision model yields lower miss rates at lower scoring cost than its prompted counterpart, supporting its use for step-level supervision during training.

\end{document}